\documentclass[sn-mathphys,Numbered]{sn-jnl}

\usepackage{graphicx}%
\usepackage{multirow}%
\usepackage{amsmath,amssymb,amsfonts}%
\usepackage{amsthm}%
\usepackage{mathrsfs}%
\usepackage[title]{appendix}%
\usepackage{xcolor}%
\usepackage{textcomp}%
\usepackage{manyfoot}%
\usepackage{booktabs}%
\usepackage{algorithm}%
\usepackage{algorithmicx}%
\usepackage{algpseudocode}%
\usepackage{listings}%

\usepackage{hyperref}   
\hypersetup{
	colorlinks=true,
	linkcolor=blue,
	citecolor=blue,
	urlcolor=blue,
}
\usepackage{kotex}
\usepackage{mathtools}
\usepackage{lipsum}
\usepackage{diagbox}
\usepackage{indentfirst}
\DeclareMathOperator*{\argmin}{argmin}
\usepackage{tabularray}

\theoremstyle{thmstyleone}%
\theoremstyle{thmstyletwo}%

\theoremstyle{thmstylethree}%

\begin{document}

\title[Article Title]{Multi-objective Generative Design Framework and Realization for Quasi-serial Manipulator: Considering Kinematic and Dynamic Performance}


\author[1]{Sumin Lee}\email{smlee1996@kaist.ac.kr}

\author[2]{Sunwoong Yang}\email{sunwoongy@kaist.ac.kr}

\author*[2]{Namwoo Kang}\email{nwkang@kaist.ac.kr}

\affil[1]{Department of Mechanical Engineering, Korea Advanced Institute of Science and Technology, 193 Munji-ro, Yuseong-gu, Daejeon 34051, Korea}

\affil[2]{Cho Chun Shik Graduate School of Mobility, Korea Advanced Institute of Science and Technology, 193 Munji-ro, Yuseong-gu, Daejeon 34051, Korea}


\abstract{This paper proposes a framework that optimizes the linkage mechanism of the quasi-serial manipulator for target tasks. This process is explained through a case study of 2-degree-of-freedom linkage mechanisms, which significantly affect the workspace of the quasi-serial manipulator. First, a vast quasi-serial mechanism is generated with a workspace satisfying a target task and it converts it into a 3D CAD model. Then, the workspace and required torque performance of each mechanism are evaluated through kinematic and dynamic analysis. A deep learning-based surrogate model is leveraged to efficiently predict mechanisms and performance during the optimization process. After model training, a multi-objective optimization problem is formulated under the mechanical and dynamic conditions of the manipulator. The design goal of the manipulator is to recommend quasi-serial mechanisms with optimized kinematic (workspace) and dynamic (joint torque) performance that satisfies the target task. To investigate the underlying physics from the obtained Pareto solutions, various data mining techniques are performed to extract design rules that can provide practical design guidance. Finally, the manipulator was designed in detail for realization with 3D printed parts, including topology optimization. Also, the task-based optimized manipulator is verified through a payload test. Based on these results, the proposed framework has the potential for other real applications as realized cases and provides a reasonable design plan through the design rule extraction.}




\maketitle
\section{Introduction} \label{INTRO}
Designing kinematic mechanisms is an important topic covered in a wide range of areas across the industry. Depending on the application, the mechanism is classified into a passive and actuating system \cite{raghavan2004suspension, bergelin2012design, wongratanaphisan2008analysis, hassan2017modeling, deshpande2017task, wang2014constant, klimchik2016design, klimchik2017serial, klimchik2016stiffness,kim2016new, shao2016conceptual, kim2008design}. For example, the passive system can be classified as automobile suspension, compensation mechanism, and the actuating system can be classified as a robot gripper, manipulator mechanism. Passive system design is mainly a structure required for fixed external conditions. Therefore, it is relatively straightforward to intuitively define design elements for optimization. In contrast, the actuation system also needs to consider the changing external conditions due to the degree of freedom. Therefore, it is significantly challenging to intuitively define design problems for optimization. Typically used as a power source in an actuating system, the motor can only perform circular motion. However, tasks in real-world applications or industries frequently require variable workspaces. The linkage mechanism is commonly used because it is inefficient to include the task region with only circular motion. The linkage mechanism has the potential to perform various tasks because it has an attractive feature that can transmit circular motion to different path motions \cite{wang2014constant, hassan2017modeling, yim2019topology, sedlaczek2009topology, li2022fourier, heyrani2022links, kim2016new, shao2016conceptual}.

The linkage mechanism transmits the torque from the drive shaft to the end-effector. The circular motion of the drive shaft is transmitted to a different workspace by kinematic motion. Workspace means space that the end-effector can reach. Torque transmission means the continual torque transfer from the drive shaft to the end-effector and is highly related to the payload. The linkage mechanism is characterized by having a significantly sensitive workspace according to the change in link length. In addition, since the mechanical changes caused by this significantly impact torque transmission, designing an appropriate mechanism remains a challenging problem. Existing design strategies to detour these problems use the method of conceiving a mechanism that satisfies the workspace and selecting an actuator that provides the required torque performance. However, these methods have significant problems in terms of optimization. First, in many cases, linkage mechanisms of the workspace that can perform tasks are not unique if there is enough design space. Therefore, the linkage mechanism optimized with the required actuator torque was not considered. Secondly, if there is no actual linkage to match the target workspace, the optimization iteration can fail to converge. Finally, the workspace and torque transmission are highly sensitive to changes in linkage length \cite{tsai2004kinematic, lee1999generalized}. Therefore, simultaneously optimizing workspace and torque transmission requires significant feedback cycles or wide-range thresholds \cite{barnawal2017evaluation, safoutin1998classification, schutze2003support}.



Numerous studies have been conducted to address these chronic problems \cite{kim2007automatic, yim2019topology, han2017topology, yu2020simultaneous, sedlaczek2009topology, yan2001kinematic, li2022fourier, heyrani2022links, deshpande2019computational, yim2021big}. They mainly used the workspace (kinematic) as an object function in optimization problems, which can be categorized into physical-based and data-based methods. The physics-based method is to design linkages that satisfy of performance objective function numerically as the target path or holding force \cite{kim2007automatic, yim2019topology, han2017topology, yu2020simultaneous, sedlaczek2009topology, yan2001kinematic, li2022fourier}. \citeauthor{yu2020simultaneous} presented a study to synthesize linkages that can track the target path with high precision, assuming a virtual zero-length spring \cite{yu2020simultaneous}. This approach has the advantage of obtaining high-precision linkages for physically feasible target paths; however, there may be some noise present or convergence issues in artificially created workspaces. Moreover, since only the target path was considered as an object function, there is an obvious limitation in that it does not take into account the torque transmission required to actuate the mechanism. \citeauthor{hassan2017modeling} also proposed a study that optimizes the structure of the gripper for holding force that has a linkage structure \cite{hassan2017modeling}. Dynamic models for maximizing grip were numerically calculated and optimized. However, there also exists a limitation that does not consider the workspace of the gripper. This physical-based methodology results in a unique solution according to constructs in a forward approach. However, in general cases, the linkages that can perform specific tasks may not be unique, so another approach is required to find different solutions that satisfy the predefined tasks. In this regard, a data-driven methodology has been proposed \cite{deshpande2019computational, yim2021big, heyrani2022links, deshpande2021image}. This strategy utilizes artificial intelligence or big data to generate a large number of linkages for the target workspace, overcoming the limitations of physics-based methods. Then, it recommends multiple candidates of linkages that meet the kinematic conditions within the numerous datasets. However, the computational cost required to collect vast amounts of data is demanding. In addition, the calculation is based on defining the kinematic motion relationship between inputs and outputs to account for dynamic conditions, and it is essential to consider physical factors such as mass and inertia. However, \citeauthor{heyrani2022links} solved the problem by assuming a rigid body link, ignoring mass and inertia, and the absence of these parts faced a complex problem in considering dynamic conditions \cite{heyrani2022links}. Therefore, to solve the optimization problem, another problem can be derived. For example, real applications that require more than one objective function could require several additional optimization steps.

In this paper, a 4-bar linkage mechanism was used to design a quasi-serial manipulator with higher structural robustness and positional precision than a serial manipulator in a high payload environment \cite{klimchik2016design, klimchik2017serial, klimchik2016stiffness}. To obtain structural advantages, task-based mechanism design is essential but existing approaches require a demanding feedback cycle and a computational cost \cite{barnawal2017evaluation, safoutin1998classification, schutze2003support}. 
In addition, it is essential to consider dynamic performance to produce realistic designs that can be applied to real applications. However, existing methodologies face several problems. Firstly, the limitation is the physics-based modeling methods, which have difficulty considering the mass and inertia of different parts for detailed design. Secondly, kinematic and dynamic performance should be considered simultaneously to optimize the task-based manipulator. Therefore, it can be effective to construct a surrogate model by generating multiple models. To address the issues at hand, we utilized a surrogate modeling approach based on machine learning (ML) and generative design. The proposed approach can replace simulation for kinematic and dynamic performance analysis. It is effective for improving design efficiency because task-based active design is enabled. Finally, the proposed novel generative framework aided by ML models can be exploited to obtain optimal mechanism design, which is scalable to other similar design problems. The main contributions of this framework can be summarized as follows:
\begin{enumerate}
\item To the authors' best knowledge, this is the first attempt to consider kinematic and dynamic performance simultaneously for quasi-serial manipulator mechanism design problems. 
\item A generalized framework for the mechanism design considering a scale factor, which can be applied to a various range size of applications as well as manipulators, is proposed.
\item To overcome the limitations of existing methods of optimizing kinematic performance, generative modeling is adopted to generate numerous task-based mechanisms considering kinematic workspace. Which has been used as a method to actively design mechanisms and can also be generally used in other applications.
\item To solve real engineering problems, a multi-objective function is usually required. In this study, 3 objective functions were selected to optimize the quasi-serial manipulator, and constraints were also considered to find the Pareto solution.
\item Design rules are extracted based on various data mining techniques to offer practical design guidance for the future design of 4-bar linkage mechanism for the quasi-serial manipulator.
\item After optimizing the mechanism design using NSGA-II, topology optimization was performed to optimize the dynamic performance. This was performed for the optimization that can be obtained from both mechanism design and shape design. 
\item The final design after topology optimization is further realized by 3D printer. The optimized dynamic performance is validated through a payload test. This test allows us to verify the performance of the realized manipulator with two optimizations and to evaluate the feasibility that can be used for the general problem of the proposed framework.
. 
\end{enumerate}

Section \ref{FRAMEWORK} describes the proposed framework process and its advantages. Section \ref{RESULT} presents the results of the proposed framework, and sensitivity analysis for data mining is also performed. And in Section \ref{REALIZATION}, the recommended optimal quasi-serial manipulator is manufactured and demonstrated. Finally, Section \ref{CONCLUSION} summarizes the conclusions and future works.







\newpage

\section{Framework}\label{FRAMEWORK}
\begin{figure*}[thb]
\centering
 \includegraphics[width=1\textwidth]{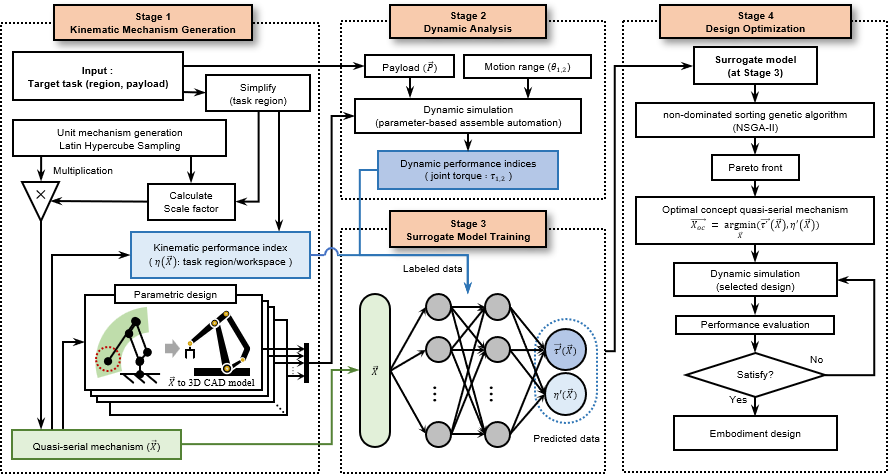}
\caption{Multi-objective generative mechanism design framework}
\label{fig1}
\vspace{-9pt}
\end{figure*}

The ML-aid optimal linkage mechanism framework proposed in this study consists of four stages, and its overall flowchart is shown in Fig. \ref{fig1}. First, it involves building an automation process aimed at generating and analyzing mechanisms capable of performing tasks through generative modeling and analysis (Stages 1 to 2). An AI-based surrogate model is then trained to predict the performance of the generated 3D CAD model. (Stage 3). Finally, the non-dominated sorting genetic algorithm-II (NSGA-II) is applied to the already trained surrogate model, recommending optimal concept mechanisms (Stage 4) \cite{deb2002fast}. The following sections are organized to demonstrate each stage of the proposed framework in detail.

\subsection{Stage 1: Kinematic mechanism generation}\label{sec:stage1}

\begin{figure}[htb]
\centering
 \includegraphics[width=0.5\textwidth]{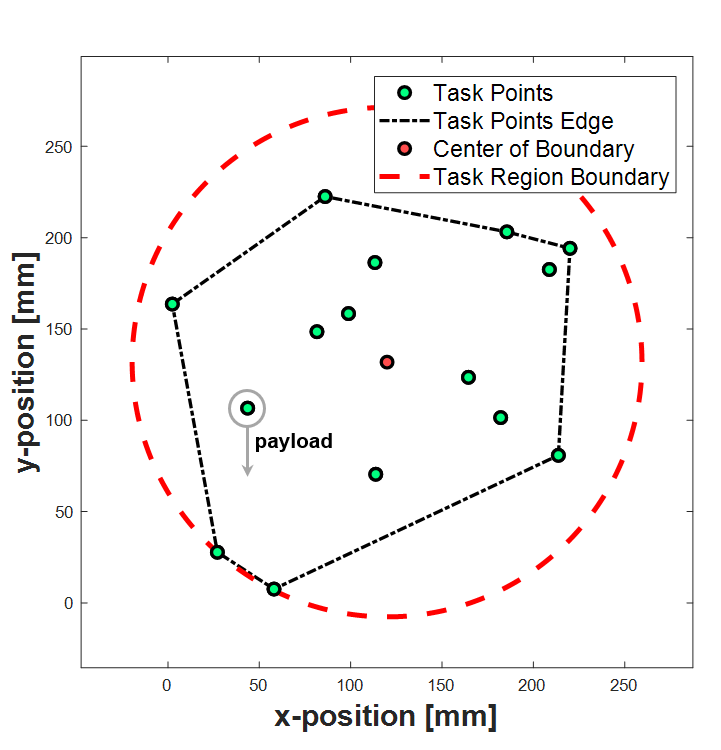}
\caption{Example of target task}
\label{fig2}
\vspace{-9pt}
\end{figure}

Stage 1 includes creating task-based quasi-serial mechanisms and converting them into 3D CAD models. As shown in Fig. \ref{fig2}, the task consists of the target region and payload. To approach a problem in general, it is important to define a task region that can contain all task points. While there are various methods to express the boundaries of the task points, this study defines a minimum circle containing all task points as a workspace to make the problem relatively easily accessible.

\begin{figure*}[htb]
\centering
 \includegraphics[width=1\textwidth]{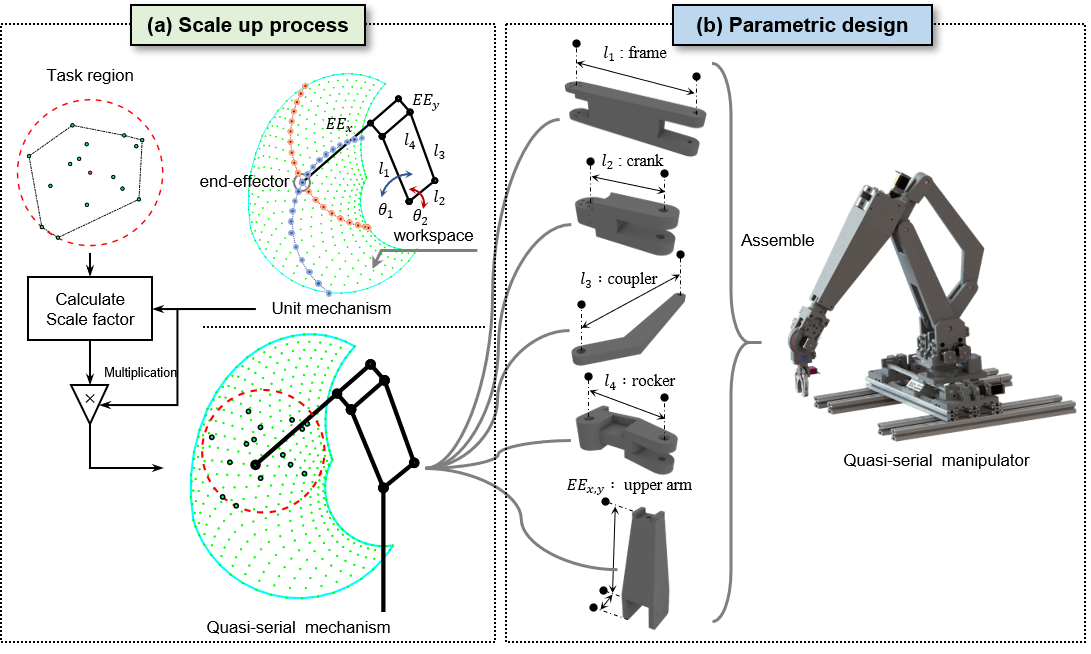}
\caption{Quasi-serial manipulator generation process}
\label{fig3}
\vspace{-9pt}
\end{figure*}

The quasi-serial mechanism is a 2-degrees of freedom (DOF) linkage consisting of a frame ($l_1$), crank ($l_2$), coupler ($l_3$), rocker ($l_4$), upper arm ($EE_x, EE_y$), and end-effector, as shown in Fig. \ref{fig3} (a). Workspace is a reachable space of the end-effector according to the motion of the frame ($\theta_1$) and crank ($\theta_2$).
The quasi-serial mechanism generation with workspaces including target regions is summarized in the following process Fig. \ref{fig3}.
The length of the unit mechanism is determined by assuming the length of the frame ($l_1$) as 1. Then, the length of each link, including $l_2$, $l_3$, $l_4$, $EE_x$, and $EE_y$, is defined as a relative ratio value to the frame length $l_1$. The various designs based on the defined link parameters are generated using Latin hypercube sampling (LHS), and the range of each design parameter is shown in Table~\ref{table_Length_range}.
When the generated designs do not satisfy the condition of crank-rocker mechanisms (Eq.~\eqref{eq_Crank-rocker}), they are filtered out.

\begin{table}[h]
\caption{Length range of the link parameter}
\label{table_Length_range}
\begin{tabular}{@{}c | c c c c c c@{}}
\toprule
 $\Vec{X} $& $l_1$ & $l_2$ & $l_3$ & $l_4$ & $EE_x$ & $EE_y$ \\
\hline
min & 1 & 0.18 & 0.8 & 0.3 & 1.0 & 0.2  \\ 
max & 1 & 0.6  & 1.3 & 0.6 & 1.4 & 0.7 \\ 
\botrule
\end{tabular}
\end{table}

\begin{equation}
l_3 + l_2 < l_1 + l_4
\label{eq_Crank-rocker}
\end{equation}

From the process above, we generated 200,000 samples and, after filtering obtained 11,080 LHS samples that satisfy unit quasi-serial mechanisms. At this point, the operating range is $45^{\circ} \le \theta_1 \le 180^{\circ} $, and $-37.5^{\circ} \le \theta_2 \le \theta_1$. The workspace of each generated unit mechanism requires a scale-up process to include the task region. The scale-up process is performed to satisfy task-dependent design issues in general. To do this, we include task regions by multiplying unit mechanisms by the appropriate scale factor. The scale factor is defined as the minimum scale for the workspace of the unit mechanism to include the task region. This figure can be fine-tuned through a threshold for design safety considerations. Therefore, the quasi-serial mechanism is calculated as the product of the unit mechanism and scale factor.

The kinematic performance of the generated quasi-serial mechanism is defined as $\eta$ = target region/workspace. The closer the value of $\eta$ is to 1, the more it means that only the target task can be included. However, since this performance only considers kinematics but not dynamics. Therefore, to take into account the dynamic performance, the generated quasi-serial mechanism is converted to a 3D CAD model by Rhinoceros 3D® software to perform dynamic analysis as in Fig. \ref{fig3} (b) \cite{mcneel-rhinoceros}.
\subsection{Stage 2: Dynamic analysis}\label{sec:stage2}
\begin{figure*}[htb]
\centering
 \includegraphics[width=1\textwidth]{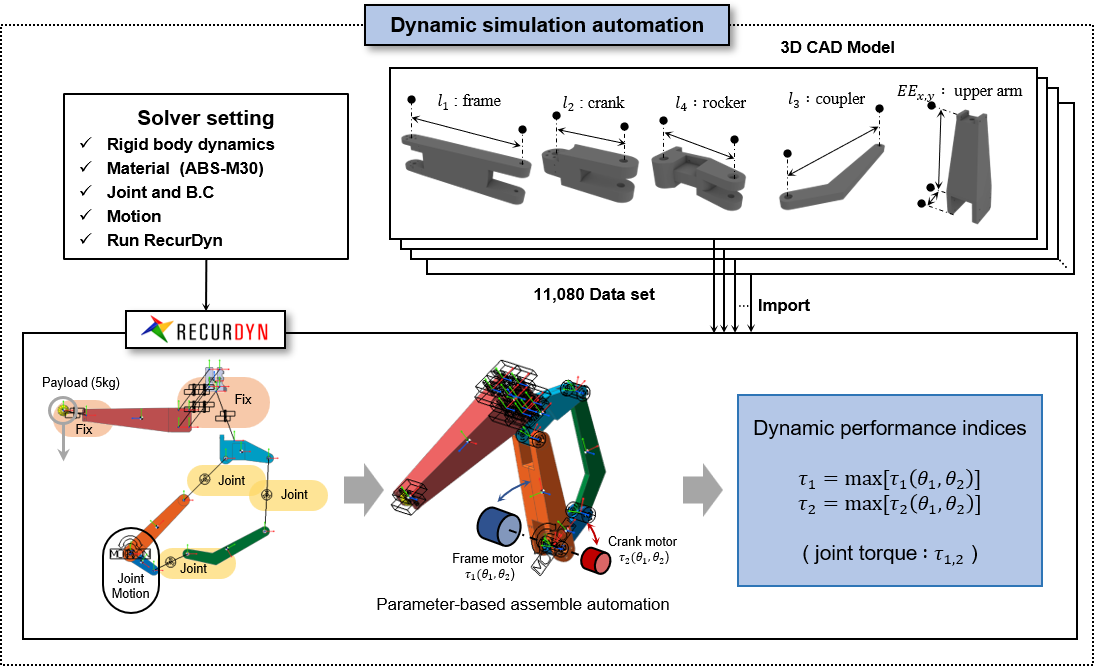}
\caption{Dynamic simulation automation process}
\label{fig4}
\vspace{-9pt}
\end{figure*}

Stage 2 includes a dynamic simulation process to obtain the dynamic performance, which will be used as a label for training the surrogate model in Stage 3. The vast amount of quasi-serial mechanisms generated in Stage 1 satisfied the target region (kinematic condition). Since Stage 1 focuses only on the kinematic performance, this section aims to focus on dynamic performance.
Therefore, all of the generated mechanisms can be a solution if the actuator performance is not considered \cite{lee2023design, kim2022bio, kim2016new, shao2016conceptual, hassan2017modeling}.
However, this approach seems significantly inappropriate from the viewpoint of efficiency and optimization. Design optimization through the surrogate model is essential to obtain the optimal mechanism in which performance is considered. The geometric performance model of a simple mechanism can be obtained by mathematical calculations \cite{hassan2017modeling, wang2014constant, bergelin2012design}. However, in a system where various mechanical components are assembled, the physics-based mathematical modeling method is significantly challenging to apply and has limitations in considering multiple performances. Therefore, training the surrogate model by performance labeling through simulation is effective. Finally, the required joint torque (actuator power:$\tau_{1,2}$, in Fig. \ref{fig4}) in the target payload (5kg) was simulated and utilized as a dynamic performance label.

\subsection{Stage 3: Surrogate model training}\label{sec:stage3}

This stage aims to train surrogate models for efficient prediction of kinematic and dynamic performances of quasi-serial mechanisms. Numerous models can be used for this purpose: to name a few, Gaussian process \cite{williams2006gaussian}, radial basis function \cite{press2007numerical}, and polynomial regression \cite{stigler1974gergonne}. Given that quasi-serial mechanisms display non-linearity in both kinematic and dynamic performances due to the geometric complexities inherent in their design, this study adopts multi-layer perceptron (MLP). It is popular in the engineering domains due to its capability of capturing complex nonlinear correlations between input and output \cite{goodfellow2016deep, jahirul2021investigation, meng2020composite, yang2023towards, yang2023inverse}: we attempt to resolve the non-linearity in the performances by using MLP model. Finally, the MLP is trained to predict both the kinematic and dynamic performance labels ($\eta, \tau_{1,2}$) obtained in Stage 2 from the quasi-serial mechanism ($\vec{X}$) obtained in Stage 1.





\subsection{Stage 4: Design optimization}\label{sec:stage4}
This stage aims to optimize kinematic and dynamic performances, which are the output of the trained MLP in Stage 3. Since the optimization process requires a number of evaluations, we leverage the already trained surrogate model to evaluate performances. Multi-objective optimization (MOO) problem is formulated since there are three objective functions in this study: $\eta$, $\tau_{1}$, and $\tau_{2}$. A well-known NSGA-II algorithm is used as the optimizer and to this end, Python package \textsc{pymoo} is utilized \cite{blank2020pymoo}. The objective function, constraints, and setting for NSGA-II are as follows:


\subsubsection{Design variables}
The six design variables are $\Vec{X}$ = [ $l_1$, $l_2$, $l_3$, $l_4$, $EE_x$, $EE_y$], where all of the elements are the quasi-serial mechanism design variables used in Section \ref{sec:stage1} (Fig. \ref{fig3}).

\subsubsection{Objective functions}
$\eta'(\Vec{X})$ is the predicted kinematic performance. We defined $\eta(\Vec{X})$ in Stage 1 as a function of $\vec{X}$. The lower this value($\eta$), the larger the workspace it has while performing the target task, and the more flexible it can be given other tasks. In Stage 3, the predicted kinetic performance by the MLP model is also defined as a function of $\vec{X}$ and used as an objective function to minimize. $\tau_1'(\Vec{X}), \tau_2'(\Vec{X})$ are the predicted joint torques required to actuate the quasi-serial manipulator, and are thus used as an objective function to minimize the actuator torques performing the same target task. Finally ($\eta', \tau_1', \tau_2'$) are selected as objective functions, all of which are minimized.



\subsubsection{Constraints}
To design a feasible quasi-serial manipulator, we have derived several constraints ($g_i(\Vec{X}), i=1, \dots, 10$).
\begin{enumerate}
\item The kinematic performance ($\eta$) of the manipulator that includes the target task should be positive. 
\begin{equation}
g_1(\Vec{X}) : \eta'(\Vec{X}) > 0
\label{eq_g1}
\end{equation}

\item The required joint torques ($\tau_1, \tau_2$) should be positive. 
\begin{align}
g_2(\Vec{X}) : \tau_1'(\Vec{X}) > 0 \\
g_3(\Vec{X}) : \tau_2'(\Vec{X}) > 0
\label{eq_g2-3}
\end{align}

\item Quasi-serial mechanism optimizations should also satisfy crank-rocker geometry Eq.~\eqref{eq_Crank-rocker}. 
\begin{equation}
g_4(\Vec{X}) : l_3 + l_2 < l_1 + l_4
\label{eq_g4}
\end{equation}

\item Lastly, $5\%$ of the designs are filtered out as outliers to bypass the singularity of the quasi-serial mechanism. We used a z-value of 1.95, assuming a normal distribution. 
\begin{align}
g_5(\Vec{X}) : \eta'(\Vec{X}) > \text{mean}[\eta(\Vec{X})] - z \cdot \text{std}[\eta(\Vec{X})] \\
g_6(\Vec{X}) : \eta'(\Vec{X}) < \text{mean}[\eta(\Vec{X})] + z \cdot \text{std}[\eta(\Vec{X})] \\
g_7(\Vec{X}) : \tau_1'(\Vec{X}) > \text{mean}[\tau_1(\Vec{X})] - z \cdot \text{std}[\tau_1(\Vec{X})] \\
g_8(\Vec{X}) : \tau_1'(\Vec{X}) < \text{mean}[\tau_1(\Vec{X})] + z \cdot \text{std}[\tau_1(\Vec{X})] \\
g_9(\Vec{X}) : \tau_2'(\Vec{X}) > \text{mean}[\tau_2(\Vec{X})] - z \cdot \text{std}[\tau_2(\Vec{X})] \\
g_{10}(\Vec{X}) : \tau_2'(\Vec{X}) < \text{mean}[\tau_2(\Vec{X})] + z \cdot \text{std}[\tau_2(\Vec{X})] 
\label{eq_g5-10}
\end{align}
\end{enumerate}

\noindent Finally, the quasi-serial manipulator optimization problem can be formulated as follows:
\begin{align}
f(\Vec{X}^*) : \displaystyle\argmin_{\vec{X}} [\eta'(\Vec{X}), \tau_1'(\Vec{X}), \tau_2'(\Vec{X})] \\
 s.t. \qquad g_i(\Vec{X}),  i = 1, \dots, 10
\label{eq_g1}
\end{align}



\section{Optimization Results}\label{RESULT}
\subsection{Dynamic analysis results}\label{sec:Dynamic_Result}
We used the dynamic simulation software RecurDyn 2023 (FunctionBay \cite{RecurDyn}), and analyzed 11,080 mechanisms for about 20 days. The summary of the dynamic simulation automation process is as follows (Fig. \ref{fig4}): 1) the manipulator parts generated in Stage 1 are imported. 2) The imported parts are automatically assembled. This is done by calculating the joint or fixed parts based on the quasi-serial mechanism parameter. 3) Each part will be produced using a 3D printer, the ABS-M30 material properties are filled. 4) The movement range of the frame and crank are input and the payload is specified. 5) To obtain the required joint torque ($\tau_{1,2}$) for the frame and crank, a Rigid Body Dynamics (RBD) analysis is performed.

\subsection{MLP model results}\label{sec:MLP_Result}
\begin{figure}[htb]
\centering
\includegraphics[width=0.5\textwidth]{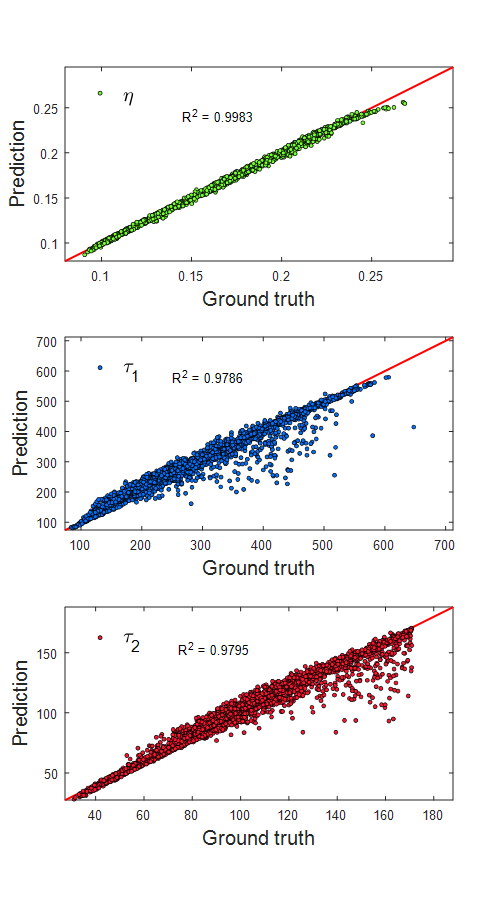}
\caption{Validation of MLP model using test data: ground truth as x-axis and prediction of MLP as y-axis}
\label{fig5}
\vspace{-9pt}
\end{figure}

Several trials and errors are undergone to find appropriate hyperparameters to avoid overfitting in the proposed application, and the final hyperparameters are as follows: the number of hidden layers is 1, the number of nodes is 100, the optimizer is Adam, max epochs is 50,000, the activation function is ReLU, and the learning rate is 0.001. Using Intel(R) Xeon(R) CPU @ 2.00GHz, training of the MLP model requires only 13.4 seconds. The presented Table \ref{table_Performance_of_the_MLP_regressor} shows the accuracy metrics of MLP predictions for both training and testing datasets: R-squared ($R^2$), mean squared error (MSE), and root mean squared error (RMSE) values were evaluated between predicted values and corresponding ground truth data to ensure an unbiased comparison between models. From the MLP model training results, the test dataset outperformed the training dataset. This can be evaluated that the learned MLP model has outstanding generalization capabilities. Also, the visual comparisons are performed in Fig. \ref{fig5}: given that most of the predictions from MLP are similar to the corresponding ground truth values, it can be concluded that MLP model has been trained successfully.

\begin{table}[h]
\caption{Performance of the MLP regressor}
\label{table_Performance_of_the_MLP_regressor}
\begin{tabular}{@{}c | c | c | c @{}}
\toprule
       & $R^2$ & MSE & RMSE  \\
\hline
Train & 0.98478 & 0.00054 & 0.02117  \\ 
Test & 0.98700 & 0.00045 & 0.01936 \\ 
\botrule
\end{tabular}
\end{table}


\subsection{Design optimization results}
\begin{figure*}[htb]
\centering
\includegraphics[width=1\textwidth]{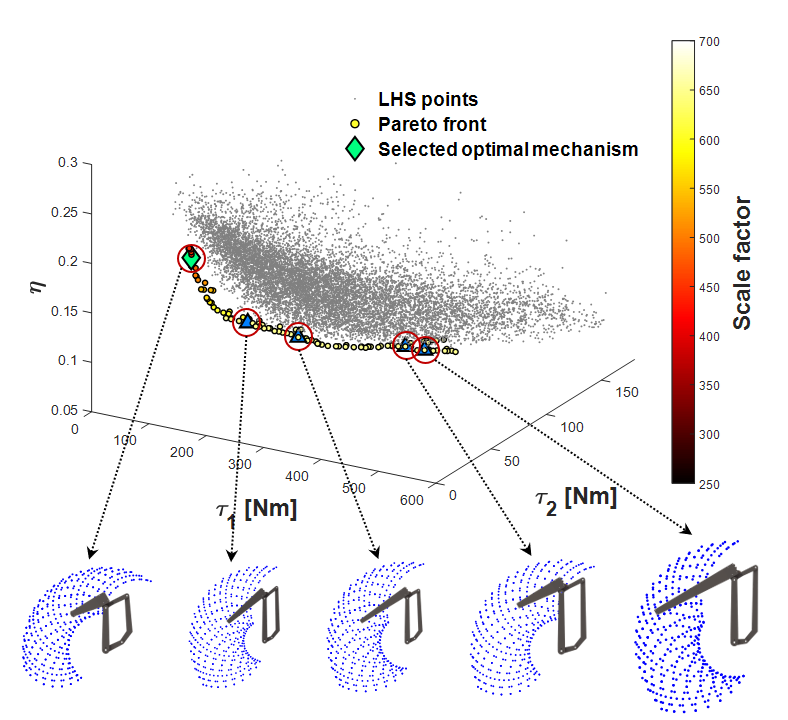}
\caption{Pareto solutions obtained by NSGA-II and configuration of 5 selected designs}
\label{fig6}
\vspace{-9pt}
\end{figure*}

Based on the trained MLP model, NSGA-II was conducted and Fig. \ref{fig6} displays the resulting Pareto front. Since the relative length for each frame was used when defining design variables in Section \ref{sec:stage1}, we now attempt to verify the effects of scale factor used to approximate the size of the manipulator. In general, the optimized solutions showed that as the scale factor increased, $\eta$ decreased and $\tau_{1,2}$ increased. Therefore, it was confirmed that the size of the manipulator generally affects the workspace and the joint torque even in the optimal solution. For a more intuitive comparison, we have visualized the manipulator and workspace for each section in Fig. \ref{fig6}. The final design to be realized by the 3D printer is selected as a diamond symbol in Fig. \ref{fig6} considering the maximum manufacturable size (the 3D printer used for this study has an output size of W × L × H: 600 mm × 350 mm × 350 mm).

\section{Design Rule Extraction}\label{sec:DM}
We conducted several additional analyses to extract design rules for quasi-serial mechanisms using already trained MLP, which enables the real-time prediction of arbitrary designs. First, in Section \ref{sec:Sobol}, a sensitivity analysis using the Sobol approach was conducted in the entire design space to determine the importance of the design parameters with respect to each objective \cite{sobol2001global, herman2017salib}. Then, in Section \ref{sec:DT}, a decision tree algorithm was applied in the design space near the Pareto solutions to extract multivariate design rules. Lastly, in Section \ref{sec:Coef}, Spearman and Pearson correlation analysis was performed for further investigation near Pareto solutions. Please refer to the work by \citet{yang2022design} for the details of the applied data mining techniques in this study.


\subsection{Sobol sensitivity analysis}\label{sec:Sobol}

\begin{figure*}[htb]
\centering
\includegraphics[width=1.0\textwidth]{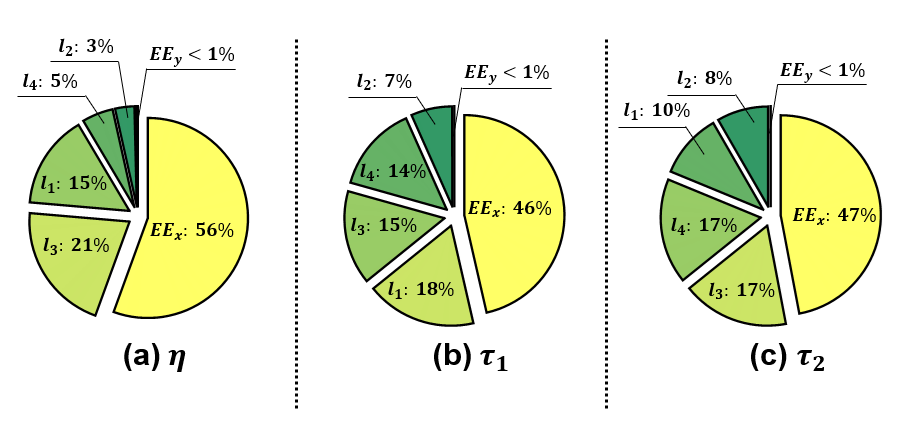}
\caption{Sobol results for each objective function.}
\label{fig7}
\vspace{-9pt}
\end{figure*}

A total of 14,336 design points were sampled on a trained MLP model using a Sobol sampler within the design space predefined: to this end, Python library \textsc{SALib} was utilized. The importance of design variables ($\vec{X}$) on the objective values ($\eta, \tau_{1,2}$) is investigated with Sobol sensitivity analysis, and the results are shown in Fig. \ref{fig7}. For $\eta$, $EE_x, l_3$, and $l_1$ are analyzed to have large contributions in that order (Fig. \ref{fig7} (a)). The reason why the $EE_x$ was rated as the most important factor is that it is the design variable closest to the end-effector. And $l_3$ was ranked second because it is related to the amount the upper arm can rotate. Finally, $l_1$ is associated with the rotational position of the upper arm and was ranked third. Since the workspace is defined as the range of the end-effector that can be reached by the rotation, it can be concluded that the objective function $\eta$ is sensitive to changes in the rotating range. On the other hand, it is relatively insensitive to $l_2, l_4,$ and $EE_y$, which mainly control position rather than rotation. To verify this more quantitatively, we mathematically derived the position of the end-effector in terms of design variables as follows.

\begin{figure}[htb]
\centering
 \includegraphics[width=0.5\textwidth]{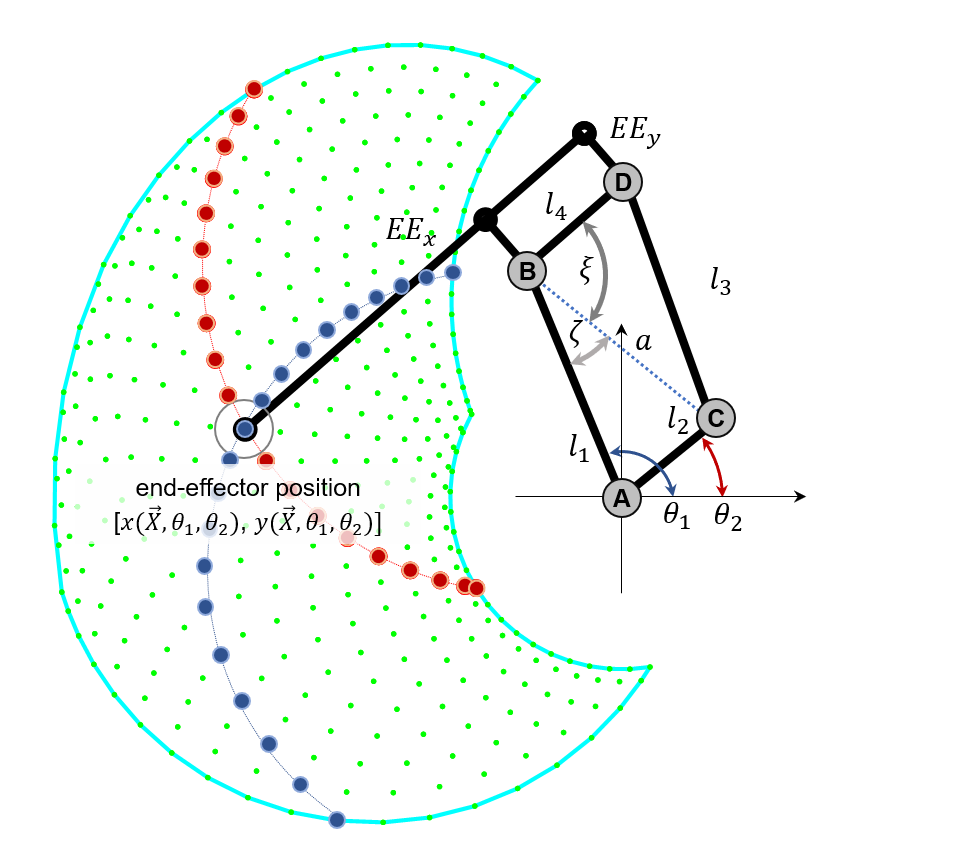}
\caption{Kinematic configuration of the quasi-serial mechanism}
\label{fig8}
\vspace{-9pt}
\end{figure}

First, the position of the frame ($l_1$) is determined by $\theta_1$ and the position of $EE_y$ is determined by the kinematic condition of the quasi-serial mechanism (see Fig. \ref{fig8}). Then, the distance $a$ between $B$ and $C$ can be obtained using Cosine law as Eq.~\eqref{eq_end-effector_1}.

\begin{equation}
a = \sqrt{l_1^2 + l_2^2 - 2 l_1 l_2 \cos(\theta_1 - \theta_2)} \\
\label{eq_end-effector_1}
\end{equation}

The angle $\angle{ABD}$ between $l_1$ and $l_4$ can be computed by the sum of $\zeta$ and $\xi$, where $\zeta$ and $\xi$ are defined as follows:


\begin{equation}
\zeta = \cos^{-1}(\frac{l_1^2 + a^2 - l_2^2}{2 l_1 a})
\label{eq_end-effector_2}
\end{equation}

\begin{equation}
\xi = \cos^{-1}(\frac{l_4^2 + a^2 - l_3^2}{2 l_4 a})
\label{eq_end-effector_3}
\end{equation}

Then, the above relation allows one to express the end-effector position [$x, y$]:

\begin{equation}
x(\Vec{X},\theta_1,\theta_2) = l_1 \cos(\theta_1) + EE_y \cos(\theta_1 + \zeta + \xi - \frac{\pi}{2}) + EE_x \cos(\theta_1 + \zeta + \xi) \\
\label{eq_end-effector_x}
\end{equation}

\begin{equation}
y(\Vec{X},\theta_1,\theta_2) = l_1 \sin(\theta_1) + EE_y \sin(\theta_1 + \zeta + \xi - \frac{\pi}{2}) + EE_x \sin(\theta_1 + \zeta + \xi) \\
\label{eq_end-effector_y}
\end{equation}

\begin{figure}[htb]
\centering
 \includegraphics[width=0.5\textwidth]{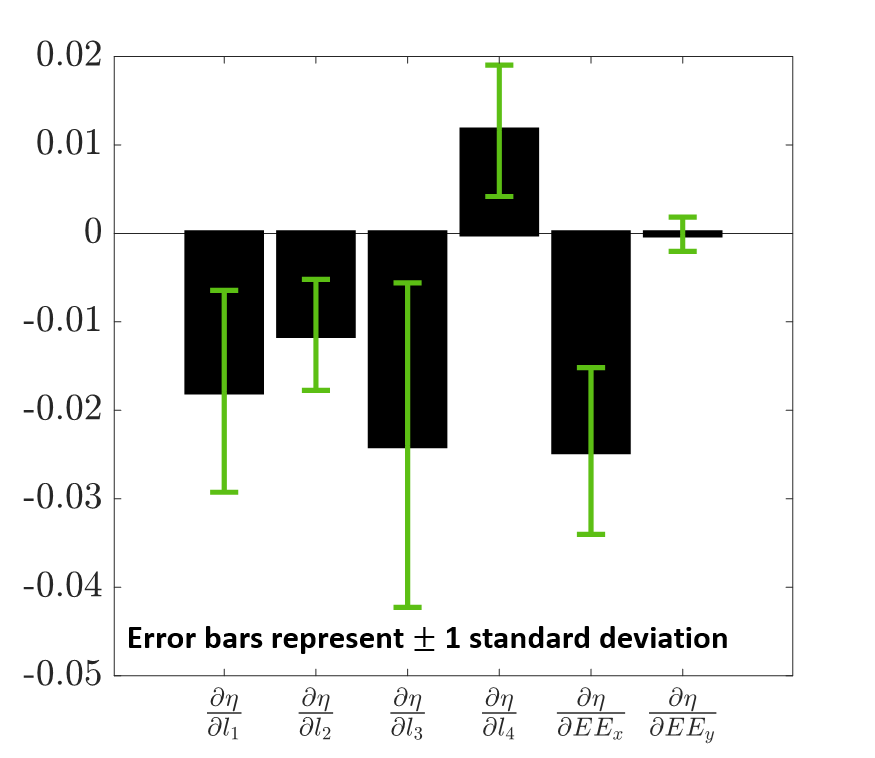}
\caption{First derivative of kinematic performance ($\eta$) with respect to design variables.}
\label{fig9}
\vspace{-9pt}
\end{figure}

The derived coordinates of the end-effector are leveraged to analyze how they change according to the design variable $\vec{X}$ of 11,080 sample designs generated in Section \ref{sec:stage1}. For this purpose, derivatives of $\eta$ with respect to six design variables are computed for all 11,080 designs and their statistics are summarized in Fig. \ref{fig9}. Interestingly, we found that the derivative values are large on the order of $EE_x, l_3$, and $l_1$, which is consistent with the results found in the Sobol analysis.


Back to the Sobol analysis, the results of $\tau_{1}$ and $\tau_{2}$ are then discussed. Note that since these torques are affected by various properties (to name a few, mass, inertia, and density) of each linkage, it is difficult to construct a mathematical model as in $\eta$. In Fig. \ref{fig7} (b), it can be found that $\tau_{1}$ has similar important design variables to $\eta$: $EE_x, l_1$, and $l_3$ reappear but the order between $l_1$, and $l_3$ changes. Again, $EE_x$, the variable closest to the workspace, was evaluated as the most influential design variable. Then, $l_1$ appeared, which is closely related to $\tau_{1}$ in that it is the rod where the torque is applied directly. The $l_3,l_4$ were ranked third and fourth without a significant difference: they constrain the coupling to 1-DOF while determining the upper arm's range of motion. Thus, it was evaluated to have a similar effect on $\tau_{1,2}$, which affects the upper arm. For $\tau_{2}$ (Fig. \ref{fig7} (c)), $EE_x$ has the most influence again. However, $l_4$ appears as the next important variable with $l_3$. As in the $\tau_{1}$, they are analyzed to have similar importance due to their role in structural robustness. These results are consistent with the properties of the quasi-serial manipulator studied previously, where it has the advantage of structural efficiency and higher rigidity with a higher payload than a serial manipulator with the same motor torque \cite{klimchik2016design, klimchik2017serial, klimchik2016stiffness}. In summary, given that the design variables extracted from the Sobol analysis are analyzed to have a reasonable underlying physics for their impact on the objective functions, it can be concluded that the trained MLP model has good enough global accuracy to be used for design optimization.


\subsection{Decision tree}\label{sec:DT}

\begin{figure*}[htb]
\centering
\includegraphics[width=1\textwidth]{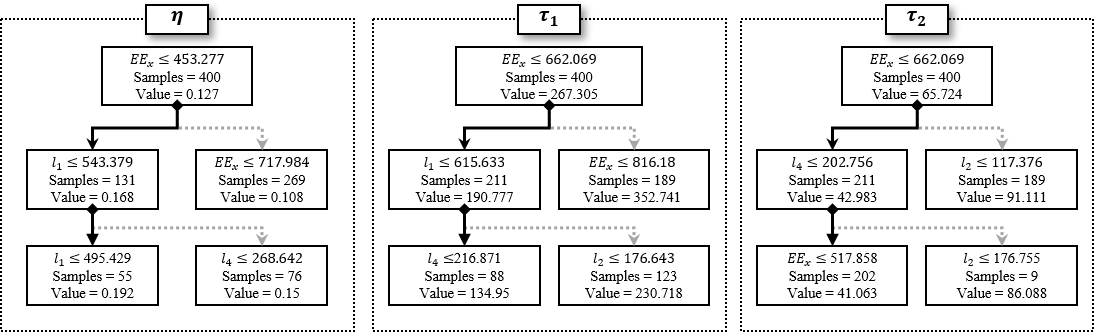}
\caption{Decision tree results for each objective function}
\label{fig10}
\vspace{-9pt}
\end{figure*}

Unlike the Sobol analysis in the previous chapter, the decision tree analysis in this chapter aims to focus on the trade-off relationships between three objective functions. For this purpose, a total of 400 sample designs were additionally sampled from the trained MLP surrogate model: 100 from the Pareto solutions and 300 from the populations of the last three generations in NSGA-II (Fig. \ref{fig14} in Appendix). Note that 300 designs were additionally considered to scrutinize the relationships between three objectives more comprehensively through exploring a design space near the optimal Pareto solutions \cite{yang2022design}. The results of the decision tree are shown in Fig. \ref{fig10}. The bold arrows indicate the direction of the optimization: for example, the arrows in the decision tree of $\eta$ point in the direction of maximizing $\eta$. For all objectives, $EE_x$ appears first, and these results can be considered consistent with the Sobol results. However, the decision trees provide further information about which direction $EE_x$ goes for each objective function. In particular, they show that there is a universal design rule that $EE_x$ should be decreased for better performance for all objective functions. It is also noteworthy that the first branch of the trees for both $\tau_1$ and $\tau_2$ is divided by the criterion $EE_x \leq 662.069$, indicating that both objectives have a similar trend with respect to the design variable $EE_x$. This can be confirmed quantitatively by the Pearson correlation coefficient between $\tau_1$ and $\tau_2$, which is calculated to be 0.95. Then, the second branches of decision trees of $\eta$ and $\tau_1$ indicate that $l_1$ should decrease for both objectives. Again, $l_1$ appeared in the Sobol results of those objectives without any information on whether it should increase or decrease. However, decision tree analysis offers the additional guideline that it should be decreased to increase $l_1$ and decrease $\tau_1$. To summarize, the results of a decision tree help engineers make decisions in the design process by showing them not only the important variables, such as Sobol analysis, but also the direction in which the variables should move for better objective values.

\subsection{Correlation coefficients}\label{sec:Coef}

\begin{figure}[htb]
\centering
 \includegraphics[width=0.5\textwidth]{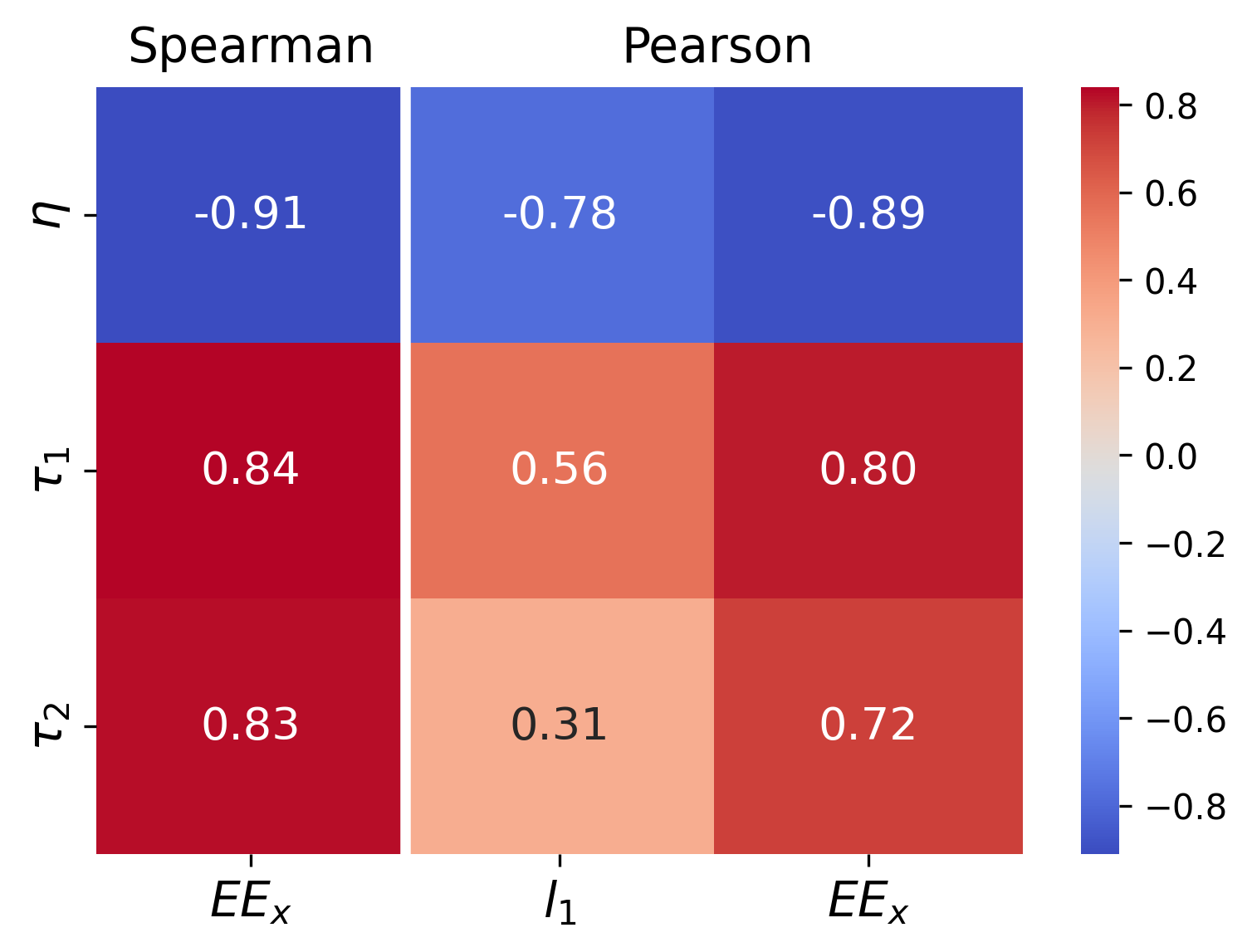}
\caption{Spearman and Pearson correlations between design variables and objective functions: only significant relations are presented}
\label{fig:corr}
\vspace{-9pt}
\end{figure}

In this section, we shift our focus to the analysis of correlation coefficients, aiming to intuitively substantiate the findings derived from the preceding decision tree analysis. This examination is to comprehensively understand the relationships between the input variables and the output variables. To achieve this, both Spearman and Pearson correlation coefficients are computed within 400 designs as in Section \ref{sec:DT}: the results are shown in Fig. \ref{fig:corr}, and only $l_1$ and $EE_x$ are found to be statistically significant with respect to outputs. 

\begin{figure*}[htb]
\centering
\includegraphics[width=1\textwidth]{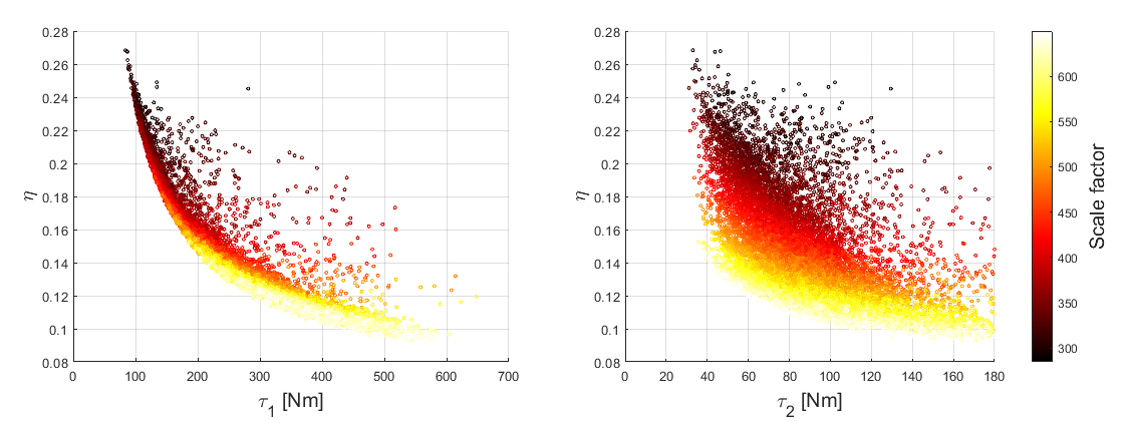}
\caption{Engineering performances of 11,080 manipulator designs colored by scale factor}
\label{fig11}
\vspace{-9pt}
\end{figure*}

Both the Spearman and Pearson correlation analysis reveal that $EE_x$ has negative correlations with $\eta$ while showing positive correlations with $\tau_1$ and $\tau_2$: these results are physically reasonable that the required torque increases as the workspace expands and the length of the upper arm increases. Also, they are in perfect agreement with the previous Sobol and decision tree analyses. Sobol denoted that $EE_x$ is always the most dominant variable for all objectives and decision tree denoted that $EE_x$ should be decreased to maximize $\eta$ and minimize $\tau_1$ and $\tau_2$. However, given that the Pearson coefficient indicates that the two variables are linearly correlated, the results of Pearson analysis additionally provide us that the variable $EE_x$ is in a linear relationship with three objectives. Moreover, Pearson coefficient between $l_1$ and $\eta$ is found to be linearly correlated with negative direction. It is also consistent with the decision tree of $\eta$: the second and third branches of it indicate that $l_1$ should decrease to maximize $\eta$. The reason for the negative linear relationships between two design variables ($l_1$ and $EE_x$) and the objective $\eta$ is that the workspace becomes wider as the length of the upper and lower arms increases: this can be verified more intuitively in Fig. \ref{fig11}. The length of $l_1$ is the scale factor, and the lengths of the other variables are determined as a ratio to $l_1$. The scale factor does not indicate the characteristics of each mechanism, but rather the overall size of the manipulator. As shown by the gradation in Fig. \ref{fig11}, there is a significant variation between manipulators with similar scale factors. Thus, it can be concluded that performance depends more on the structure of the mechanism rather than the size of the manipulator required to perform the target task. This suggests that design optimization of quasi-serial mechanisms is essential.

\section{Realizing Optimal Quasi-Serial Manipulator Design through 3D Printing}\label{REALIZATION}

\begin{figure*}[htb]
\centering
 \includegraphics[width=1\textwidth]{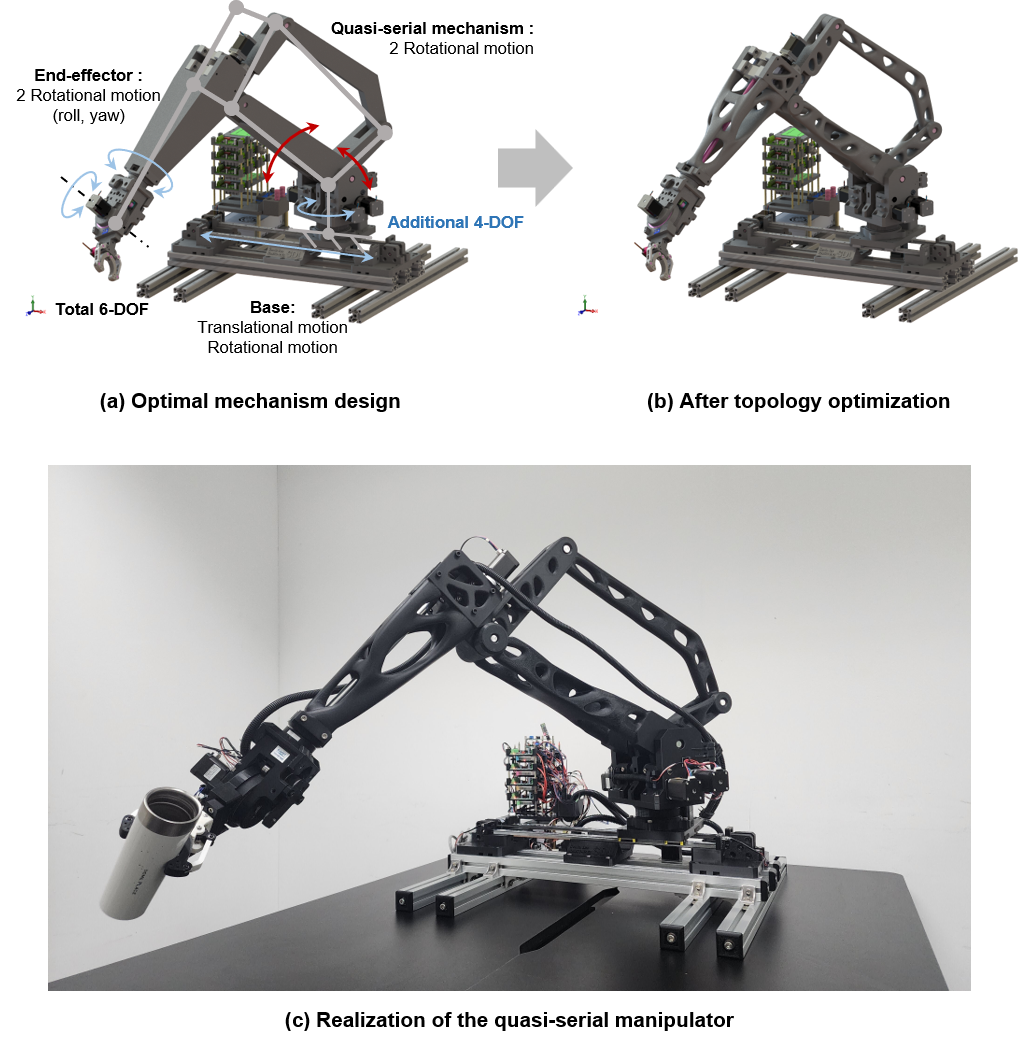}
\caption{Embodiment design and topology optimization}
\label{fig12}
\vspace{-9pt}
\end{figure*}

In this section, we have the embodiment design of the quasi-serial manipulator and include what we build and actuate it. We got 100 optional concept quasi-serial manipulator candidates using the proposed framework. We selected candidates with the appropriate size that we can manufacture the manipulator using 3D printer (Fig. \ref{fig6}. We focused on the mechanism of 2-DOF, which has a major impact on the quasi-serial manipulator to perform the target task. However, the actual manipulator usually has 6-DOF. Therefore, we designed the additional 4-DOF part required in the embodiment design step. 2-DOF was added to the base part so the manipulator could rotate and translate. The end-effector's roll, yaw DOF has been added so that the object can be grabbed at various angles and positions (Fig. \ref{fig12} (a)). To implement the feasible design, which also included the configuration of motors, gears, bearings, controllers, etc. In addition, topology optimization was performed to improve further the dynamic performance of the recommended optimization (Fig. \ref{fig12} (b)). Finally, components were assembled to obtain the actual quasi-serial manipulator (Fig. \ref{fig12} (c)).

\begin{figure*}[htb]
\centering
 \includegraphics[width=1\textwidth]{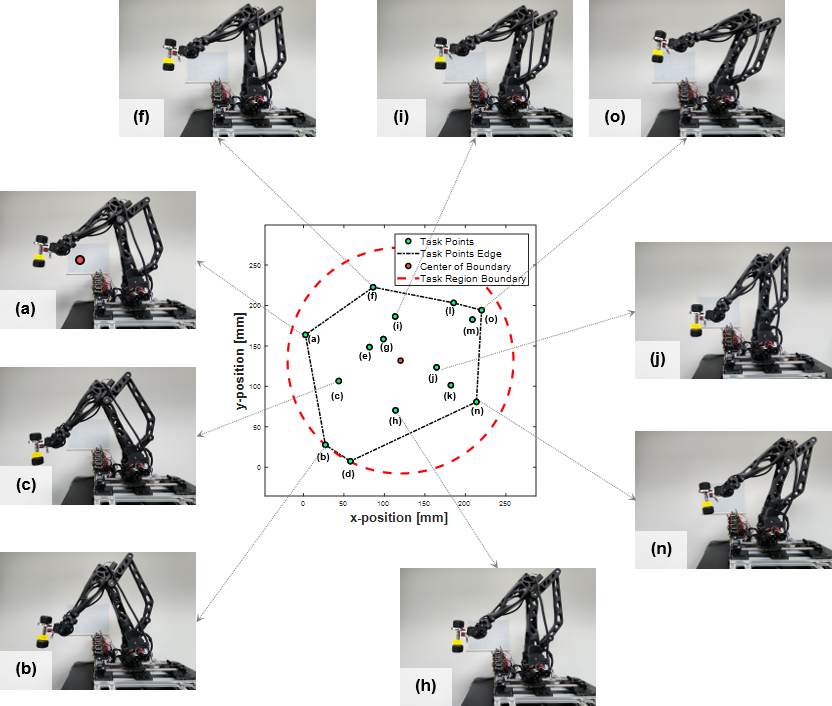}
\caption{Realized manipulator payload test actuation}
\label{fig13}
\vspace{-9pt}
\end{figure*}

The required actuator specification before and after topology optimization for payload (5 kg) is the same as in Table~\ref{table_performance_cahnge}. The motor used to achieve the required actuator specification is a NEMA17 stepper motor designed for a maximum output of 150 Nm and 100 Nm using a 50:1 gear and timing belt. A typical motor has a characteristic where output angular velocity and torque are inversely proportional. Therefore, to output the same torque as shown in Table~\ref{table_performance_cahnge}, the required torque was satisfied by adjusting the speeds of the stepper motor. Experiments were also conducted to perform the tasks shown in Figure~\ref{fig2}. The payload used in the dynamic simulation was 5 kg, but the weight of the gripper added during the embodiment design process was about 2 kg. Therefore, the 3 kg object was gripped and the task was performed (Fig. \ref{fig13}).

\begin{table}[h]
\caption{Joint torques with/without topology optimization}
\label{table_performance_cahnge}
\begin{tabular}{@{}c l c l c @{}}
\toprule
 &   $\tau_1$ [Nm] & $\tau_2$ [Nm]  \\
\hline
 Without topology optimization & 132.6011 & 37.63381 \\ 
 After topology optimization & 98.58294 & 31.4224  \\
  \hline
 Reduced ratio & 25.65 \%  & 16.50 \%  \\
\botrule
\end{tabular}
\end{table}

\section{Conclusion}\label{CONCLUSION}
This study is a data-driven design study that has never been attempted in the assembly environment of mechanisms considering kinematic and dynamic performance simultaneously, and there is a great novelty in confirming its feasibility through this study. In addition, it is not limited to the quasi-serial manipulator used as an application. We plan to explore comprehensive design issues by using them for appropriate applications for design processes that require feedback cycles with future work.

In this paper, the framework for the mechanism design that optimizes the kinematic and dynamic performance of the quasi-serial manipulator was proposed. In the mechanism generation stage, a scale-up process was performed to include the target region, and it was converted into a 3D CAD model through parametric design. During the dynamic simulation stage, the joint torque of the generated quasi-serial manipulator was obtained through analysis automation. The kinematic (workspace) and dynamic (joint torque) performances of the previous stage were used as objective functions to train MLP surrogate models and formulate optimization problems. We defined the MOO problem to optimize the kinematic and dynamic performance while generating the task-based mechanism: Pareto solutions were obtained through NSGA-II. Data mining techniques including Sobol sensitivity analysis and decision tree analysis were performed, which extracted the design rules that can offer practical design guidance for future preliminary design of quasi-serial manipulator. The design rules extraction was used to analyze the influence of each design variable. This provided an effective indicator for optimization, during the detailed design stage of realization.

Moreover, topology optimization was performed in the detailed design process for realization, and as the mass and inertia of the link decreased, the required $\tau_1, \tau_2$ also decreased. The topology-optimized manipulator was manufactured and assembled using a 3D printer and a payload test was performed, and the feasibility of the proposed framework was verified.


Although we have aimed to solve a required feedback cycle of the performance optimization problem, the proposed framework has several limitations. We approached the problem under the assumption that the motor specification meets the required point torque is continuous. However, the performance of commercial motors is usually discontinuous, and there may be a few more specifications to consider other than the torque (resolution, friction, etc). The next limitation is that only the actuator was considered to drive the link; however, in reality, components such as gearboxes, pulleys, chains, etc. can be used for power transmission. In this context, the insufficient performance of a single component can cause the performance degradation of the entire system. To address these limitations, mechanical components can be classified for real-world applications and the proposed framework can be performed using a labeled look-up table with performance values. However, this procedure requires basic data to determine design specifications, which can be time-consuming when investigating multiple parts, such as a manipulator.

\bmhead{Acknowledgments} 


\paragraph{Funding.}
This work was supported by the National Research Foundation of Korea grant (2018R1A5A7025409) and the Ministry of Science and ICT of Korea grant (No.2022-0-00969, No.2022-0-00986).
\section*{Declarations}
The authors declare that they have no known competing financial interests or personal relationships that could influence the work reported herein.

\clearpage
\begin{appendices}

\section{Design candidates used for data mining}\label{secA1}

\begin{figure}[htb]
\centering
 \includegraphics[width=0.5\textwidth]{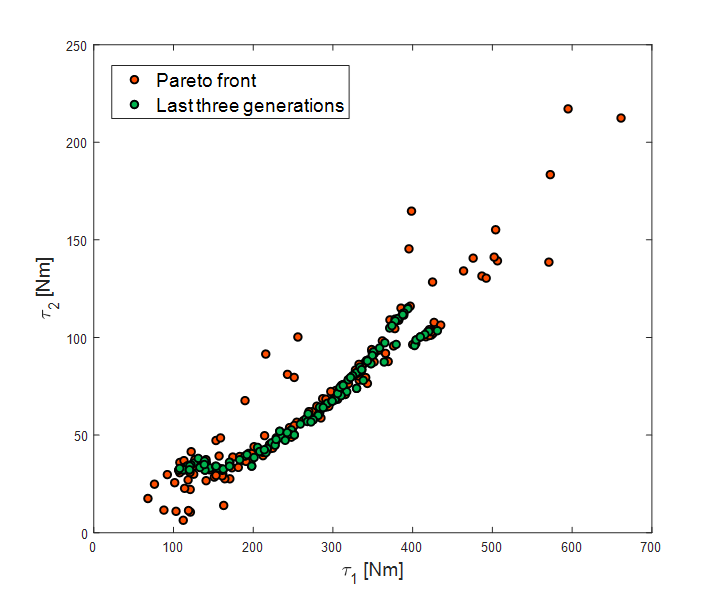}
\caption{Extracted 400 designs used for data mining in Section \ref{sec:DT} and \ref{sec:Coef}}
\label{fig14}
\vspace{-9pt}
\end{figure}



\end{appendices}


\bibliography{sn-bibliography}


\begin{thebibliography}{45}
\ifx \bisbn   \undefined \def \bisbn  #1{ISBN #1}\fi
\ifx \binits  \undefined \def \binits#1{#1}\fi
\ifx \bauthor  \undefined \def \bauthor#1{#1}\fi
\ifx \batitle  \undefined \def \batitle#1{#1}\fi
\ifx \bjtitle  \undefined \def \bjtitle#1{#1}\fi
\ifx \bvolume  \undefined \def \bvolume#1{\textbf{#1}}\fi
\ifx \byear  \undefined \def \byear#1{#1}\fi
\ifx \bissue  \undefined \def \bissue#1{#1}\fi
\ifx \bfpage  \undefined \def \bfpage#1{#1}\fi
\ifx \blpage  \undefined \def \blpage #1{#1}\fi
\ifx \burl  \undefined \def \burl#1{\textsf{#1}}\fi
\ifx \doiurl  \undefined \def \doiurl#1{\url{https://doi.org/#1}}\fi
\ifx \betal  \undefined \def \betal{\textit{et al.}}\fi
\ifx \binstitute  \undefined \def \binstitute#1{#1}\fi
\ifx \binstitutionaled  \undefined \def \binstitutionaled#1{#1}\fi
\ifx \bctitle  \undefined \def \bctitle#1{#1}\fi
\ifx \beditor  \undefined \def \beditor#1{#1}\fi
\ifx \bpublisher  \undefined \def \bpublisher#1{#1}\fi
\ifx \bbtitle  \undefined \def \bbtitle#1{#1}\fi
\ifx \bedition  \undefined \def \bedition#1{#1}\fi
\ifx \bseriesno  \undefined \def \bseriesno#1{#1}\fi
\ifx \blocation  \undefined \def \blocation#1{#1}\fi
\ifx \bsertitle  \undefined \def \bsertitle#1{#1}\fi
\ifx \bsnm \undefined \def \bsnm#1{#1}\fi
\ifx \bsuffix \undefined \def \bsuffix#1{#1}\fi
\ifx \bparticle \undefined \def \bparticle#1{#1}\fi
\ifx \barticle \undefined \def \barticle#1{#1}\fi
\bibcommenthead
\ifx \bconfdate \undefined \def \bconfdate #1{#1}\fi
\ifx \botherref \undefined \def \botherref #1{#1}\fi
\ifx \url \undefined \def \url#1{\textsf{#1}}\fi
\ifx \bchapter \undefined \def \bchapter#1{#1}\fi
\ifx \bbook \undefined \def \bbook#1{#1}\fi
\ifx \bcomment \undefined \def \bcomment#1{#1}\fi
\ifx \oauthor \undefined \def \oauthor#1{#1}\fi
\ifx \citeauthoryear \undefined \def \citeauthoryear#1{#1}\fi
\ifx \endbibitem  \undefined \def \endbibitem {}\fi
\ifx \bconflocation  \undefined \def \bconflocation#1{#1}\fi
\ifx \arxivurl  \undefined \def \arxivurl#1{\textsf{#1}}\fi
\csname PreBibitemsHook\endcsname

\bibitem[\protect\citeauthoryear{Raghavan}{2004}]{raghavan2004suspension}
\begin{barticle}
\bauthor{\bsnm{Raghavan}, \binits{M.}}:
\batitle{Suspension design for linear toe curves: a case study in mechanism synthesis}.
\bjtitle{J. Mech. Des.}
\bvolume{126}(\bissue{2}),
\bfpage{278}--\blpage{282}
(\byear{2004})
\end{barticle}
\endbibitem

\bibitem[\protect\citeauthoryear{Bergelin and Voglewede}{2012}]{bergelin2012design}
\begin{botherref}
\oauthor{\bsnm{Bergelin}, \binits{B.J.}},
\oauthor{\bsnm{Voglewede}, \binits{P.A.}}:
Design of an active ankle-foot prosthesis utilizing a four-bar mechanism
(2012)
\end{botherref}
\endbibitem

\bibitem[\protect\citeauthoryear{Wongratanaphisan and Cole}{2008}]{wongratanaphisan2008analysis}
\begin{botherref}
\oauthor{\bsnm{Wongratanaphisan}, \binits{T.}},
\oauthor{\bsnm{Cole}, \binits{M.O.}}:
Analysis of a gravity compensated four-bar linkage mechanism with linear spring suspension
(2008)
\end{botherref}
\endbibitem

\bibitem[\protect\citeauthoryear{Hassan and Abomoharam}{2017}]{hassan2017modeling}
\begin{barticle}
\bauthor{\bsnm{Hassan}, \binits{A.}},
\bauthor{\bsnm{Abomoharam}, \binits{M.}}:
\batitle{Modeling and design optimization of a robot gripper mechanism}.
\bjtitle{Robotics and Computer-Integrated Manufacturing}
\bvolume{46},
\bfpage{94}--\blpage{103}
(\byear{2017})
\end{barticle}
\endbibitem

\bibitem[\protect\citeauthoryear{Deshpande and Purwar}{2017}]{deshpande2017task}
\begin{barticle}
\bauthor{\bsnm{Deshpande}, \binits{S.}},
\bauthor{\bsnm{Purwar}, \binits{A.}}:
\batitle{A task-driven approach to optimal synthesis of planar four-bar linkages for extended burmester problem}.
\bjtitle{Journal of Mechanisms and Robotics}
\bvolume{9}(\bissue{6}),
\bfpage{061005}
(\byear{2017})
\end{barticle}
\endbibitem

\bibitem[\protect\citeauthoryear{Wang and Lan}{2014}]{wang2014constant}
\begin{barticle}
\bauthor{\bsnm{Wang}, \binits{J.-Y.}},
\bauthor{\bsnm{Lan}, \binits{C.-C.}}:
\batitle{A constant-force compliant gripper for handling objects of various sizes}.
\bjtitle{Journal of Mechanical Design}
\bvolume{136}(\bissue{7}),
\bfpage{071008}
(\byear{2014})
\end{barticle}
\endbibitem

\bibitem[\protect\citeauthoryear{Klimchik et~al.}{2016}]{klimchik2016design}
\begin{barticle}
\bauthor{\bsnm{Klimchik}, \binits{A.}},
\bauthor{\bsnm{Magid}, \binits{E.}},
\bauthor{\bsnm{Pashkevich}, \binits{A.}}:
\batitle{Design of experiments for elastostatic calibration of heavy industrial robots with kinematic parallelogram and gravity compensator}.
\bjtitle{IFAC-PapersOnLine}
\bvolume{49}(\bissue{12}),
\bfpage{967}--\blpage{972}
(\byear{2016})
\end{barticle}
\endbibitem

\bibitem[\protect\citeauthoryear{Klimchik and Pashkevich}{2017}]{klimchik2017serial}
\begin{barticle}
\bauthor{\bsnm{Klimchik}, \binits{A.}},
\bauthor{\bsnm{Pashkevich}, \binits{A.}}:
\batitle{Serial vs. quasi-serial manipulators: Comparison analysis of elasto-static behaviors}.
\bjtitle{Mechanism and Machine Theory}
\bvolume{107},
\bfpage{46}--\blpage{70}
(\byear{2017})
\end{barticle}
\endbibitem

\bibitem[\protect\citeauthoryear{Klimchik et~al.}{2016}]{klimchik2016stiffness}
\begin{bchapter}
\bauthor{\bsnm{Klimchik}, \binits{A.}},
\bauthor{\bsnm{Magid}, \binits{E.}},
\bauthor{\bsnm{Caro}, \binits{S.}},
\bauthor{\bsnm{Waiyakan}, \binits{K.}},
\bauthor{\bsnm{Pashkevich}, \binits{A.}}:
\bctitle{Stiffness of serial and quasi-serial manipulators: comparison analysis}.
In: \bbtitle{2016 International Conference on Mechanical, System and Control Engineering (ICMSC 2016)}
(\byear{2016})
\end{bchapter}
\endbibitem

\bibitem[\protect\citeauthoryear{Kim et~al.}{2016}]{kim2016new}
\begin{barticle}
\bauthor{\bsnm{Kim}, \binits{J.-W.}},
\bauthor{\bsnm{Seo}, \binits{T.}},
\bauthor{\bsnm{Kim}, \binits{J.}}:
\batitle{A new design methodology for four-bar linkage mechanisms based on derivations of coupler curve}.
\bjtitle{Mechanism and Machine Theory}
\bvolume{100},
\bfpage{138}--\blpage{154}
(\byear{2016})
\end{barticle}
\endbibitem

\bibitem[\protect\citeauthoryear{Shao et~al.}{2016}]{shao2016conceptual}
\begin{barticle}
\bauthor{\bsnm{Shao}, \binits{Y.}},
\bauthor{\bsnm{Xiang}, \binits{Z.}},
\bauthor{\bsnm{Liu}, \binits{H.}},
\bauthor{\bsnm{Li}, \binits{L.}}:
\batitle{Conceptual design and dimensional synthesis of cam-linkage mechanisms for gait rehabilitation}.
\bjtitle{Mechanism and Machine Theory}
\bvolume{104},
\bfpage{31}--\blpage{42}
(\byear{2016})
\end{barticle}
\endbibitem

\bibitem[\protect\citeauthoryear{Kim et~al.}{2008}]{kim2008design}
\begin{bchapter}
\bauthor{\bsnm{Kim}, \binits{S.-K.}},
\bauthor{\bsnm{Shin}, \binits{W.-H.}},
\bauthor{\bsnm{Ko}, \binits{S.-Y.}},
\bauthor{\bsnm{Kim}, \binits{J.}},
\bauthor{\bsnm{Kwon}, \binits{D.-S.}}:
\bctitle{Design of a compact 5-dof surgical robot of a spherical mechanism: Cures}.
In: \bbtitle{2008 IEEE/ASME International Conference on Advanced Intelligent Mechatronics},
pp. \bfpage{990}--\blpage{995}
(\byear{2008}).
\bcomment{IEEE}
\end{bchapter}
\endbibitem

\bibitem[\protect\citeauthoryear{Yim et~al.}{2019}]{yim2019topology}
\begin{barticle}
\bauthor{\bsnm{Yim}, \binits{N.H.}},
\bauthor{\bsnm{Kang}, \binits{S.W.}},
\bauthor{\bsnm{Kim}, \binits{Y.Y.}}:
\batitle{Topology optimization of planar gear-linkage mechanisms}.
\bjtitle{Journal of Mechanical Design}
\bvolume{141}(\bissue{3}),
\bfpage{032301}
(\byear{2019})
\end{barticle}
\endbibitem

\bibitem[\protect\citeauthoryear{Sedlaczek and Eberhard}{2009}]{sedlaczek2009topology}
\begin{botherref}
\oauthor{\bsnm{Sedlaczek}, \binits{K.}},
\oauthor{\bsnm{Eberhard}, \binits{P.}}:
Topology optimization of large motion rigid body mechanisms with nonlinear kinematics
(2009)
\end{botherref}
\endbibitem

\bibitem[\protect\citeauthoryear{Li et~al.}{2022}]{li2022fourier}
\begin{barticle}
\bauthor{\bsnm{Li}, \binits{X.}},
\bauthor{\bsnm{Lv}, \binits{H.}},
\bauthor{\bsnm{Zhao}, \binits{P.}},
\bauthor{\bsnm{Lu}, \binits{Q.}}:
\batitle{A fourier approach to kinematic acquisition of geometric constraints of planar motion for practical mechanism design}.
\bjtitle{Journal of Mechanical Design}
\bvolume{144}(\bissue{12}),
\bfpage{123302}
(\byear{2022})
\end{barticle}
\endbibitem

\bibitem[\protect\citeauthoryear{Heyrani~Nobari et~al.}{2022}]{heyrani2022links}
\begin{bchapter}
\bauthor{\bsnm{Heyrani~Nobari}, \binits{A.}},
\bauthor{\bsnm{Srivastava}, \binits{A.}},
\bauthor{\bsnm{Gutfreund}, \binits{D.}},
\bauthor{\bsnm{Ahmed}, \binits{F.}}:
\bctitle{Links: A dataset of a hundred million planar linkage mechanisms for data-driven kinematic design}.
In: \bbtitle{International Design Engineering Technical Conferences and Computers and Information in Engineering Conference},
vol. \bseriesno{86229},
pp. \bfpage{03}--\blpage{03013}
(\byear{2022}).
\bcomment{American Society of Mechanical Engineers}
\end{bchapter}
\endbibitem

\bibitem[\protect\citeauthoryear{Tsai and Lai}{2004}]{tsai2004kinematic}
\begin{barticle}
\bauthor{\bsnm{Tsai}, \binits{M.-J.}},
\bauthor{\bsnm{Lai}, \binits{T.-H.}}:
\batitle{Kinematic sensitivity analysis of linkage with joint clearance based on transmission quality}.
\bjtitle{Mechanism and Machine Theory}
\bvolume{39}(\bissue{11}),
\bfpage{1189}--\blpage{1206}
(\byear{2004})
\end{barticle}
\endbibitem

\bibitem[\protect\citeauthoryear{Lee et~al.}{1999}]{lee1999generalized}
\begin{barticle}
\bauthor{\bsnm{Lee}, \binits{M.-Y.}},
\bauthor{\bsnm{Erdman}, \binits{A.}},
\bauthor{\bsnm{Faik}, \binits{S.}}:
\batitle{A generalized performance sensitivity synthesis methodology for four-bar mechanisms}.
\bjtitle{Mechanism and machine theory}
\bvolume{34}(\bissue{7}),
\bfpage{1127}--\blpage{1139}
(\byear{1999})
\end{barticle}
\endbibitem

\bibitem[\protect\citeauthoryear{Barnawal et~al.}{2017}]{barnawal2017evaluation}
\begin{barticle}
\bauthor{\bsnm{Barnawal}, \binits{P.}},
\bauthor{\bsnm{Dorneich}, \binits{M.C.}},
\bauthor{\bsnm{Frank}, \binits{M.C.}},
\bauthor{\bsnm{Peters}, \binits{F.}}:
\batitle{Evaluation of design feedback modality in design for manufacturability}.
\bjtitle{Journal of Mechanical Design}
\bvolume{139}(\bissue{9}),
\bfpage{094503}
(\byear{2017})
\end{barticle}
\endbibitem

\bibitem[\protect\citeauthoryear{Safoutin and Smith}{1998}]{safoutin1998classification}
\begin{bchapter}
\bauthor{\bsnm{Safoutin}, \binits{M.J.}},
\bauthor{\bsnm{Smith}, \binits{R.P.}}:
\bctitle{Classification of iteration in engineering design processes}.
In: \bbtitle{International Design Engineering Technical Conferences and Computers and Information in Engineering Conference},
vol. \bseriesno{80333},
pp. \bfpage{003}--\blpage{03012}
(\byear{1998}).
\bcomment{American Society of Mechanical Engineers}
\end{bchapter}
\endbibitem

\bibitem[\protect\citeauthoryear{Sch{\"u}tze et~al.}{2003}]{schutze2003support}
\begin{barticle}
\bauthor{\bsnm{Sch{\"u}tze}, \binits{M.}},
\bauthor{\bsnm{Sachse}, \binits{P.}},
\bauthor{\bsnm{R{\"o}mer}, \binits{A.}}:
\batitle{Support value of sketching in the design process}.
\bjtitle{Research in Engineering Design}
\bvolume{14},
\bfpage{89}--\blpage{97}
(\byear{2003})
\end{barticle}
\endbibitem

\bibitem[\protect\citeauthoryear{Kim et~al.}{2007}]{kim2007automatic}
\begin{botherref}
\oauthor{\bsnm{Kim}, \binits{Y.Y.}},
\oauthor{\bsnm{Jang}, \binits{G.-W.}},
\oauthor{\bsnm{Park}, \binits{J.H.}},
\oauthor{\bsnm{Hyun}, \binits{J.S.}},
\oauthor{\bsnm{Nam}, \binits{S.J.}}:
Automatic synthesis of a planar linkage mechanism with revolute joints by using spring-connected rigid block models
(2007)
\end{botherref}
\endbibitem

\bibitem[\protect\citeauthoryear{Han et~al.}{2017}]{han2017topology}
\begin{barticle}
\bauthor{\bsnm{Han}, \binits{S.M.}},
\bauthor{\bsnm{In~Kim}, \binits{S.}},
\bauthor{\bsnm{Kim}, \binits{Y.Y.}}:
\batitle{Topology optimization of planar linkage mechanisms for path generation without prescribed timing}.
\bjtitle{Structural and Multidisciplinary Optimization}
\bvolume{56},
\bfpage{501}--\blpage{517}
(\byear{2017})
\end{barticle}
\endbibitem

\bibitem[\protect\citeauthoryear{Yu et~al.}{2020}]{yu2020simultaneous}
\begin{barticle}
\bauthor{\bsnm{Yu}, \binits{J.}},
\bauthor{\bsnm{Han}, \binits{S.M.}},
\bauthor{\bsnm{Kim}, \binits{Y.Y.}}:
\batitle{Simultaneous shape and topology optimization of planar linkage mechanisms based on the spring-connected rigid block model}.
\bjtitle{Journal of Mechanical Design}
\bvolume{142}(\bissue{1}),
\bfpage{011401}
(\byear{2020})
\end{barticle}
\endbibitem

\bibitem[\protect\citeauthoryear{Yan and Soong}{2001}]{yan2001kinematic}
\begin{barticle}
\bauthor{\bsnm{Yan}, \binits{H.-S.}},
\bauthor{\bsnm{Soong}, \binits{R.-C.}}:
\batitle{Kinematic and dynamic design of four-bar linkages by links counterweighing with variable input speed}.
\bjtitle{Mechanism and machine theory}
\bvolume{36}(\bissue{9}),
\bfpage{1051}--\blpage{1071}
(\byear{2001})
\end{barticle}
\endbibitem

\bibitem[\protect\citeauthoryear{Deshpande and Purwar}{2019}]{deshpande2019computational}
\begin{barticle}
\bauthor{\bsnm{Deshpande}, \binits{S.}},
\bauthor{\bsnm{Purwar}, \binits{A.}}:
\batitle{Computational creativity via assisted variational synthesis of mechanisms using deep generative models}.
\bjtitle{Journal of Mechanical Design}
\bvolume{141}(\bissue{12}),
\bfpage{121402}
(\byear{2019})
\end{barticle}
\endbibitem

\bibitem[\protect\citeauthoryear{Yim et~al.}{2021}]{yim2021big}
\begin{barticle}
\bauthor{\bsnm{Yim}, \binits{N.H.}},
\bauthor{\bsnm{Lee}, \binits{J.}},
\bauthor{\bsnm{Kim}, \binits{J.}},
\bauthor{\bsnm{Kim}, \binits{Y.Y.}}:
\batitle{Big data approach for the simultaneous determination of the topology and end-effector location of a planar linkage mechanism}.
\bjtitle{Mechanism and Machine Theory}
\bvolume{163},
\bfpage{104375}
(\byear{2021})
\end{barticle}
\endbibitem

\bibitem[\protect\citeauthoryear{Deshpande and Purwar}{2021}]{deshpande2021image}
\begin{barticle}
\bauthor{\bsnm{Deshpande}, \binits{S.}},
\bauthor{\bsnm{Purwar}, \binits{A.}}:
\batitle{An image-based approach to variational path synthesis of linkages}.
\bjtitle{Journal of Computing and Information Science in Engineering}
\bvolume{21}(\bissue{2}),
\bfpage{021005}
(\byear{2021})
\end{barticle}
\endbibitem

\bibitem[\protect\citeauthoryear{Deb et~al.}{2002}]{deb2002fast}
\begin{barticle}
\bauthor{\bsnm{Deb}, \binits{K.}},
\bauthor{\bsnm{Pratap}, \binits{A.}},
\bauthor{\bsnm{Agarwal}, \binits{S.}},
\bauthor{\bsnm{Meyarivan}, \binits{T.}}:
\batitle{A fast and elitist multiobjective genetic algorithm: Nsga-ii}.
\bjtitle{IEEE transactions on evolutionary computation}
\bvolume{6}(\bissue{2}),
\bfpage{182}--\blpage{197}
(\byear{2002})
\end{barticle}
\endbibitem

\bibitem[\protect\citeauthoryear{McNeel et~al.}{2020}]{mcneel-rhinoceros}
\begin{botherref}
\oauthor{\bsnm{McNeel}, \binits{R.}}, et al.:
Rhinoceros 3d, version 7.0.
Robert McNeel \& Associates, Seattle, WA
(2020)
\end{botherref}
\endbibitem

\bibitem[\protect\citeauthoryear{Lee et~al.}{2023}]{lee2023design}
\begin{barticle}
\bauthor{\bsnm{Lee}, \binits{S.}},
\bauthor{\bsnm{Choi}, \binits{S.}},
\bauthor{\bsnm{Ko}, \binits{C.}},
\bauthor{\bsnm{Kim}, \binits{T.}},
\bauthor{\bsnm{Kong}, \binits{K.}}:
\batitle{Design and control of the compact cable-driven series elastic actuator module in soft wearable robot for ankle assistance}.
\bjtitle{International Journal of Control, Automation and Systems}
\bvolume{21}(\bissue{5}),
\bfpage{1624}--\blpage{1633}
(\byear{2023})
\end{barticle}
\endbibitem

\bibitem[\protect\citeauthoryear{Kim et~al.}{2022}]{kim2022bio}
\begin{barticle}
\bauthor{\bsnm{Kim}, \binits{J.}},
\bauthor{\bsnm{Kim}, \binits{T.}},
\bauthor{\bsnm{Ko}, \binits{C.}},
\bauthor{\bsnm{Lee}, \binits{S.}},
\bauthor{\bsnm{Kong}, \binits{K.}}:
\batitle{Bio-inspired cable-driven knee orthosis for tibiofemoral joint load distribution}.
\bjtitle{IFAC-PapersOnLine}
\bvolume{55}(\bissue{27}),
\bfpage{430}--\blpage{435}
(\byear{2022})
\end{barticle}
\endbibitem

\bibitem[\protect\citeauthoryear{Williams and Rasmussen}{2006}]{williams2006gaussian}
\begin{bbook}
\bauthor{\bsnm{Williams}, \binits{C.K.}},
\bauthor{\bsnm{Rasmussen}, \binits{C.E.}}:
\bbtitle{Gaussian Processes for Machine Learning}
vol. \bseriesno{2}.
\bpublisher{MIT press},
\blocation{Cambridge, MA}
(\byear{2006})
\end{bbook}
\endbibitem

\bibitem[\protect\citeauthoryear{Press}{2007}]{press2007numerical}
\begin{bbook}
\bauthor{\bsnm{Press}, \binits{W.H.}}:
\bbtitle{Numerical Recipes 3rd Edition: The Art of Scientific Computing}.
\bpublisher{Cambridge university press},
\blocation{Cambridge, England}
(\byear{2007})
\end{bbook}
\endbibitem

\bibitem[\protect\citeauthoryear{Stigler}{1974}]{stigler1974gergonne}
\begin{barticle}
\bauthor{\bsnm{Stigler}, \binits{S.M.}}:
\batitle{Gergonne's 1815 paper on the design and analysis of polynomial regression experiments}.
\bjtitle{Historia Mathematica}
\bvolume{1}(\bissue{4}),
\bfpage{431}--\blpage{439}
(\byear{1974})
\end{barticle}
\endbibitem

\bibitem[\protect\citeauthoryear{Goodfellow et~al.}{2016}]{goodfellow2016deep}
\begin{bbook}
\bauthor{\bsnm{Goodfellow}, \binits{I.}},
\bauthor{\bsnm{Bengio}, \binits{Y.}},
\bauthor{\bsnm{Courville}, \binits{A.}}:
\bbtitle{Deep Learning}.
\bpublisher{MIT press},
\blocation{Cambridge, MA}
(\byear{2016})
\end{bbook}
\endbibitem

\bibitem[\protect\citeauthoryear{Jahirul et~al.}{2021}]{jahirul2021investigation}
\begin{barticle}
\bauthor{\bsnm{Jahirul}, \binits{M.}},
\bauthor{\bsnm{Rasul}, \binits{M.}},
\bauthor{\bsnm{Brown}, \binits{R.}},
\bauthor{\bsnm{Senadeera}, \binits{W.}},
\bauthor{\bsnm{Hosen}, \binits{M.}},
\bauthor{\bsnm{Haque}, \binits{R.}},
\bauthor{\bsnm{Saha}, \binits{S.}},
\bauthor{\bsnm{Mahlia}, \binits{T.}}:
\batitle{Investigation of correlation between chemical composition and properties of biodiesel using principal component analysis (pca) and artificial neural network (ann)}.
\bjtitle{Renewable energy}
\bvolume{168},
\bfpage{632}--\blpage{646}
(\byear{2021})
\end{barticle}
\endbibitem

\bibitem[\protect\citeauthoryear{Meng and Karniadakis}{2020}]{meng2020composite}
\begin{barticle}
\bauthor{\bsnm{Meng}, \binits{X.}},
\bauthor{\bsnm{Karniadakis}, \binits{G.E.}}:
\batitle{A composite neural network that learns from multi-fidelity data: Application to function approximation and inverse pde problems}.
\bjtitle{Journal of Computational Physics}
\bvolume{401},
\bfpage{109020}
(\byear{2020})
\end{barticle}
\endbibitem

\bibitem[\protect\citeauthoryear{Yang and Yee}{2023}]{yang2023towards}
\begin{botherref}
\oauthor{\bsnm{Yang}, \binits{S.}},
\oauthor{\bsnm{Yee}, \binits{K.}}:
Towards quantifying calibrated uncertainty via deep ensembles in multi-output regression task.
arXiv preprint arXiv:2303.16210
(2023)
\end{botherref}
\endbibitem

\bibitem[\protect\citeauthoryear{Yang et~al.}{2023}]{yang2023inverse}
\begin{barticle}
\bauthor{\bsnm{Yang}, \binits{S.}},
\bauthor{\bsnm{Lee}, \binits{S.}},
\bauthor{\bsnm{Yee}, \binits{K.}}:
\batitle{Inverse design optimization framework via a two-step deep learning approach: application to a wind turbine airfoil}.
\bjtitle{Engineering with Computers}
\bvolume{39}(\bissue{3}),
\bfpage{2239}--\blpage{2255}
(\byear{2023})
\end{barticle}
\endbibitem

\bibitem[\protect\citeauthoryear{Blank and Deb}{2020}]{blank2020pymoo}
\begin{barticle}
\bauthor{\bsnm{Blank}, \binits{J.}},
\bauthor{\bsnm{Deb}, \binits{K.}}:
\batitle{Pymoo: Multi-objective optimization in python}.
\bjtitle{IEEE Access}
\bvolume{8},
\bfpage{89497}--\blpage{89509}
(\byear{2020})
\end{barticle}
\endbibitem

\bibitem[\protect\citeauthoryear{FunctionBay}{2023}]{RecurDyn}
\begin{botherref}
\oauthor{\bsnm{FunctionBay}}:
RecurDyn, Version 2023.
(2023)
\end{botherref}
\endbibitem

\bibitem[\protect\citeauthoryear{Sobol}{2001}]{sobol2001global}
\begin{barticle}
\bauthor{\bsnm{Sobol}, \binits{I.M.}}:
\batitle{Global sensitivity indices for nonlinear mathematical models and their monte carlo estimates}.
\bjtitle{Mathematics and computers in simulation}
\bvolume{55}(\bissue{1-3}),
\bfpage{271}--\blpage{280}
(\byear{2001})
\end{barticle}
\endbibitem

\bibitem[\protect\citeauthoryear{Herman and Usher}{2017}]{herman2017salib}
\begin{barticle}
\bauthor{\bsnm{Herman}, \binits{J.}},
\bauthor{\bsnm{Usher}, \binits{W.}}:
\batitle{Salib: An open-source python library for sensitivity analysis}.
\bjtitle{Journal of Open Source Software}
\bvolume{2}(\bissue{9}),
\bfpage{97}
(\byear{2017})
\end{barticle}
\endbibitem

\bibitem[\protect\citeauthoryear{Yang and Yee}{2022}]{yang2022design}
\begin{barticle}
\bauthor{\bsnm{Yang}, \binits{S.}},
\bauthor{\bsnm{Yee}, \binits{K.}}:
\batitle{Design rule extraction using multi-fidelity surrogate model for unmanned combat aerial vehicles}.
\bjtitle{Journal of Aircraft}
\bvolume{59}(\bissue{4}),
\bfpage{977}--\blpage{991}
(\byear{2022})
\end{barticle}
\endbibitem

\end{thebibliography}

\end{document}